\documentclass[conference]{IEEEtran}
\IEEEoverridecommandlockouts

\usepackage{cite}
\usepackage{amsmath,amssymb,amsfonts}
\usepackage{graphicx}
\usepackage{textcomp}
\usepackage{xcolor}
\usepackage{booktabs}
\usepackage{array}
\usepackage{multirow}
\usepackage{url}
\usepackage{hyperref}
\hypersetup{colorlinks=false}

\def\BibTeX{{\rm B\kern-.05em{\sc i\kern-.025em b}\kern-.08em
    T\kern-.1667em\lower.7ex\hbox{E}\kern-.125emX}}

\begin{document}
 
\title{Benchmarking Classical and Transformer-Based Models for Document Sensitivity Classification}
 
\author{
  \IEEEauthorblockN{Aleesha Zainab$^{1,2*}$, Muhammad Ahmed Khalid$^{1}$,
  Faheem Ullah Khan$^{1}$, Asifullah Khan$^{1,2,3}$}
  \IEEEauthorblockA{$^{1}$\textit{Pattern Recognition Lab, DCIS, PIEAS, Nilore, Islamabad, 45650, Pakistan}}
  \IEEEauthorblockA{$^{2}$\textit{PIEAS Artificial Intelligence Center (PAIC), PIEAS, Nilore, Islamabad, 45650, Pakistan}}
  \IEEEauthorblockA{$^{3}$\textit{Deep Learning Lab, Center for Mathematical Sciences, PIEAS, Nilore, Islamabad, 45650, Pakistan}}
  \IEEEauthorblockA{$^{*}$Corresponding author: aleeshazainab660@gmail.com}
}
 
\maketitle

\begin{abstract}
Automatic sensitivity classification of organizational documents is
a critical yet underserved problem, where the consequences of
misclassification range from regulatory violations to security
breaches. While AI-based approaches offer a scalable alternative to
manual review, their reliability depends fundamentally on the
integrity of training data. A pervasive but underreported problem in
this domain is label leakage: residual classification markers
embedded within document bodies that allow models to exploit surface
shortcuts rather than learning genuine content-based sensitivity
signals, producing performance estimates that are inflated and
unreliable. This paper addresses this problem by introducing
\textit{Strategic 16K}, a carefully constructed, leakage-controlled
corpus of 16,000 diplomatic cables sourced from the WikiLeaks Public
Library of US Diplomacy (PlusD), and presents a systematic benchmark
evaluating six model architectures spanning classical machine
learning and transformer-based approaches. We document an extended
leakage removal protocol that identifies and eliminates three
categories of residual classification markers embedded within
document bodies. On the clean benchmark, BERT achieves the strongest
performance (Accuracy\,=\,89.14\%, F1\,=\,89.33\%), followed by
ELECTRA (Accuracy\,=\,88.57\%, F1\,=\,88.90\%). Among classical
models, TF-IDF with Logistic Regression achieves the strongest
performance at significantly lower computational cost. These results
constitute the first fully reproducible sensitivity classification
benchmark constructed under explicit leakage-controlled conditions
from WikiLeaks PlusD.
\end{abstract}


\section{Introduction}

A single misclassified document can lead to a data breach, a
regulatory penalty, or a national security incident. In most
organizations, the classification of a document as sensitive is
still determined at the document level, manually. Before it can be
routed, stored, and transmitted, trained reviewers must read and
label each document manually, which is time consuming, inconsistent
across reviewers and cannot scale with the volume of documents
today's organizations produce. While automated sensitivity
classification with NLP provides a principled scalable solution, the
utility of this approach is highly dependent on whether the trained
models capture actual sensitivity signals or are simply capturing
artefacts of how the training data was prepared.

There is a significance in this difference that is greater than
might be supposed. \textbf{Label leakage} is a well-known and often
overlooked document sensitivity classification problem, where
explicit sensitivity markers are found in the body text of training
documents, such as by being embedded in inline classification codes,
classification phrases, and distribution notices. The ground-truth
label is directly stored in a machine-readable way on these
artefacts. Such a trained classifier does not need to know what
makes a document sensitive, it just needs to see when a
classification marker is present. The model gets very high
cross-validation scores but when the model is used on documents that
have been correctly sanitised, there it fails, which is the
operational condition.

This problem is not merely theoretical. Prior studies using the
WikiLeaks Public Library of US Diplomacy (PlusD) as a training
corpus for sensitivity classification generally do not disclose
explicit leakage removal procedures. Their reported scores cannot be
interpreted as evidence of genuine sensitivity understanding without
knowing what artefacts were contained within the training data; and
cannot be fairly compared across studies or reproduced in controlled
conditions.

This paper will directly address these limitations. We build
\textit{Strategic 16K}, a 16K-document benchmark from WikiLeaks
PlusD, using a well-documented leakage removal protocol, and
perform the first controlled cross-family evaluation under identical
experimental conditions on six model architectures. The
contributions of this work are:

\begin{enumerate}
  \item The first reproducible sensitivity classification benchmark
  from WikiLeaks PlusD, with fully documented leakage removal and
  sensitive-focused sampling protocols.

  \item An extended leakage removal protocol targeting three
  categories of residual classification markers embedded within
  document bodies that prior studies have not addressed.

  \item A systematic cross-family benchmark comparing classical
  TF-IDF classifiers and transformer-based models under identical
  experimental conditions on a common cleaned corpus.

  \item A quantified efficiency analysis identifying TF-IDF with
  Logistic Regression as a strong practical baseline, at a fraction
  of the computational cost of transformer fine-tuning.
\end{enumerate}

\section{Related Work}

\subsection{Traditional Machine Learning Approaches}

However, with the help of TF-IDF and bag-of-words representations,
classical machine learning methods have set solid baselines for text
classification in various domains. Ahmad et al.~\cite{ahmad2025}
tested SVM, Random Forest, k-Nearest Neighbours and Naive Bayes for
document categorisation, and found SVM to be 97\% accurate, proving
that well-regularized linear models work well with lexically rich
text. Joachims~\cite{joachims1998} gave theoretical foundations for
SVM-based text categorisation by showing that the maximum-margin
formulation with sparse TF-IDF representations has good
generalisation bounds in high-dimensional feature spaces. Although
these methods are efficient, they can only compare words at the
surface level and do not account for the bidirectional relationships
that are often key in deciding whether a document is sensitive or
not.

\subsection{Deep Learning Approaches}

Alzhrani et al.~\cite{alzhrani2019} built a CNN-based system for
sensitive text detection on the dataset of the WikiLeaks diplomatic
cables with an F1 score of 0.91, segmenting the long documents into
shorter paragraphs to redirect the focus toward locally sensitive
passages. This represents the first application of deep learning to
WikiLeaks PlusD sensitivity classification. In a previous work,
Hart et al.~\cite{hart2011} have presented the possibility of using
machine learning for Data Loss Prevention (DLP) activities, arguing
that automatic classifiers can be used to detect sensitive
information in organisational documents. A critical limitation shared by both studies is that they do not explicitly describe any procedure for identifying or removing
potential label leakage from the WikiLeaks corpus. Consequently,
it is difficult to determine whether the reported performance
reflects genuine semantic sensitivity understanding or, at least in
part, the exploitation of residual classification artefacts.

\subsection{Transformer-Based Models}

A bidirectional transformer pre-trained on the masked language
modelling task, BERT~\cite{devlin2019} changed the paradigm in text
classification. BERT represents tokens with context, using
self-attention at all layers to model the left and right context of
a token simultaneously, which cannot be achieved with TF-IDF or
recurrent models. Petrolini et al.~\cite{petrolini2022} fine-tuned
BERT to detect sensitive data automatically, and obtained
F1\,=\,0.95 on a privacy-sensitive classification task. Kowsari et
al.~\cite{kowsari2019} conducted a survey of text classification
algorithms and found that transformer models have shown consistently
and significantly better performance than classical and recurrent
models.Saritha and Kumar~\cite{saritha2025} presented a review of
AI-based techniques for sensitive data protection, highlighting the
use of transformer models, convolutional neural networks (CNNs),
recurrent neural networks (RNNs), autoencoders, and metaheuristic
optimization methods for detecting and preventing data leakage.
However, the reviewed studies did not address explicit leakage
control during dataset construction.

\subsection{Summary of Research Gaps}

Representative previous studies on document and text classification
are summarized in Table~\ref{tab:related}. The studies come from
various environments and data sets and are added to give a context
to the model families and evaluation methods for sensitivity
classification. The present study is motivated by four gaps observed
across all the reviewed work: (1) No leakage removal protocol is
explicitly disclosed in prior studies that used WikiLeaks PlusD;
(2) previous benchmarks are challenging to reproduce because they
do not specify preprocessing pipelines and dataset splits; (3) No
controlled cross-family comparison was conducted on a common cleaned
corpus; and (4) No work quantifies the efficiency tradeoff between
transformer and classical models in this domain.

\begin{table}[h]
\caption{Representative Prior Work on Text and Sensitivity Classification}
\label{tab:related}
\centering
\renewcommand{\arraystretch}{1.2}
\begin{tabular}{llcc}
\toprule
\textbf{Study} & \textbf{Method} & \textbf{Score} & \textbf{Leakage Ctrl} \\
\midrule
Ahmad et al.~\cite{ahmad2025}         & SVM, NB     & Acc=0.97   & No \\
Joachims~\cite{joachims1998}          & SVM         & High Acc.  & No \\
Alzhrani et al.~\cite{alzhrani2019}   & CNN         & F1=0.91    & No \\
Hart et al.~\cite{hart2011} & ML-based DLP & FNR<3\%, FDR<1\% & No \\
Petrolini et al.~\cite{petrolini2022} & BERT        & F1=0.95    & No \\
\textbf{This work} & \textbf{6 models} & \textbf{F1=0.893} & \textbf{Yes} \\
\bottomrule
\end{tabular}
\end{table}

\section{Dataset and Preprocessing}

\subsection{Source: WikiLeaks Public Library of US Diplomacy}

The WikiLeaks Public Library of US Diplomacy (PlusD) is the
largest publicly available collection of real government documents
with official sensitivity labels, containing 251,287 diplomatic
cables produced by US embassies worldwide. The seven official
classification labels -- UNCLASSIFIED, CONFIDENTIAL, LIMITED
OFFICIAL USE, SECRET, UNCLASSIFIED//FOR OFFICIAL USE ONLY (FOUO),
CONFIDENTIAL//NOFORN, and SECRET//NOFORN -- were affixed to each
cable by trained government officials at the time of its writing.
These are real government classification decisions using formal
protocols and not crowd-sourced annotations, meaning that this
dataset is uniquely reliable as a source of real ground-truth
sensitivity labels.

There was no pre-built version of this data available. Every
document was collected from the WikiLeaks PlusD API via a custom
automated extraction pipeline. The pipeline repeatedly processed
each classification category, fetched HTML pages of cables,
extracted the document content and original classification labels,
and stripped away any leftover HTML artefacts. The pipeline
deduplicated more than 100,000 unique documents, which were stored
in structured CSV format.

\subsection{Binary Label Mapping}

The seven original labels were reduced to a binary schema, which
represents the basic access-restriction decision. UNCLASSIFIED and
UNCLASSIFIED//FOUO were both classified as Non-Sensitive since they
do not have formal access restrictions. All the other labels
(LIMITED OFFICIAL USE, CONFIDENTIAL, CONFIDENTIAL//NOFORN, SECRET,
and SECRET//NOFORN) were assigned a formal access restriction of
Sensitive, since all apply some degree of formal restriction. This
mapping and the resulting distribution of the corpus are shown in
Table~\ref{tab:dataset}.

\subsection{Leakage Removal Protocol}

One of the most important challenges during the preprocessing of
WikiLeaks PlusD cables is the removal of remaining classification
artefacts that are embedded in document bodies. Unless eliminated,
these can enable classifiers to use surface shortcuts instead of
learning real sensitivity signals. Three types of leakage were found
and eliminated:

\begin{itemize}
  \item \textbf{Inline paragraph markers:} Single-letter sensitivity
  codes \texttt{(C)}, \texttt{(S)}, \texttt{(U)} and \texttt{(SBU)}
  inserted at the beginning of each paragraph of a specific cable
  indicating the paragraph classification.

  \item \textbf{Repeated classification phrases:} Natural-language
  phrases like \textit{``this cable is classified SECRET''} in the
  body of the document and not in the header or metadata fields.

  \item \textbf{Distribution notices:} Repetitive boilerplate
  strings such as \textit{``Sensitive But Unclassified, Not for
  Internet Distribution''} repeated throughout document bodies.
\end{itemize}

All recognized patterns have been eliminated during the extended
cleaning pass. Sensitive terms reflecting the classification were
left within natural, semi-natural sentence contexts. For example, a
cable discussing nuclear confidentiality agreements was not modified.
This is the most crucial difference between artefact removal and
content preservation, which underlie the construction of a benchmark
that forces models to learn from real content.

To illustrate the protocol, three representative before-and-after
examples are provided: (1) \texttt{(C) The ambassador confirmed...}
becomes \texttt{The ambassador confirmed...}; (2) \texttt{This cable
is classified SECRET. The meeting...} becomes \texttt{The meeting...};
and (3) \texttt{Sensitive But Unclassified, Not for Internet
Distribution. Officials agreed...} becomes
\texttt{Officials agreed...}.

Following the cleaning pass, the corpus was additionally examined
for letter-spaced classification banners, a pattern documented in
raw PlusD cables in which classification labels appear as spaced
characters (e.g., \texttt{S\,E\,C\,R\,E\,T} as a standalone line).
No such patterns were identified in the sampled subset. The three
categories described above therefore represent the complete set of
leakage artefacts present in Strategic 16K, and the corpus is
considered fully leakage-controlled with respect to the document
sample it comprises.

\begin{figure}[h]
  \centering
  \includegraphics[width=\columnwidth]{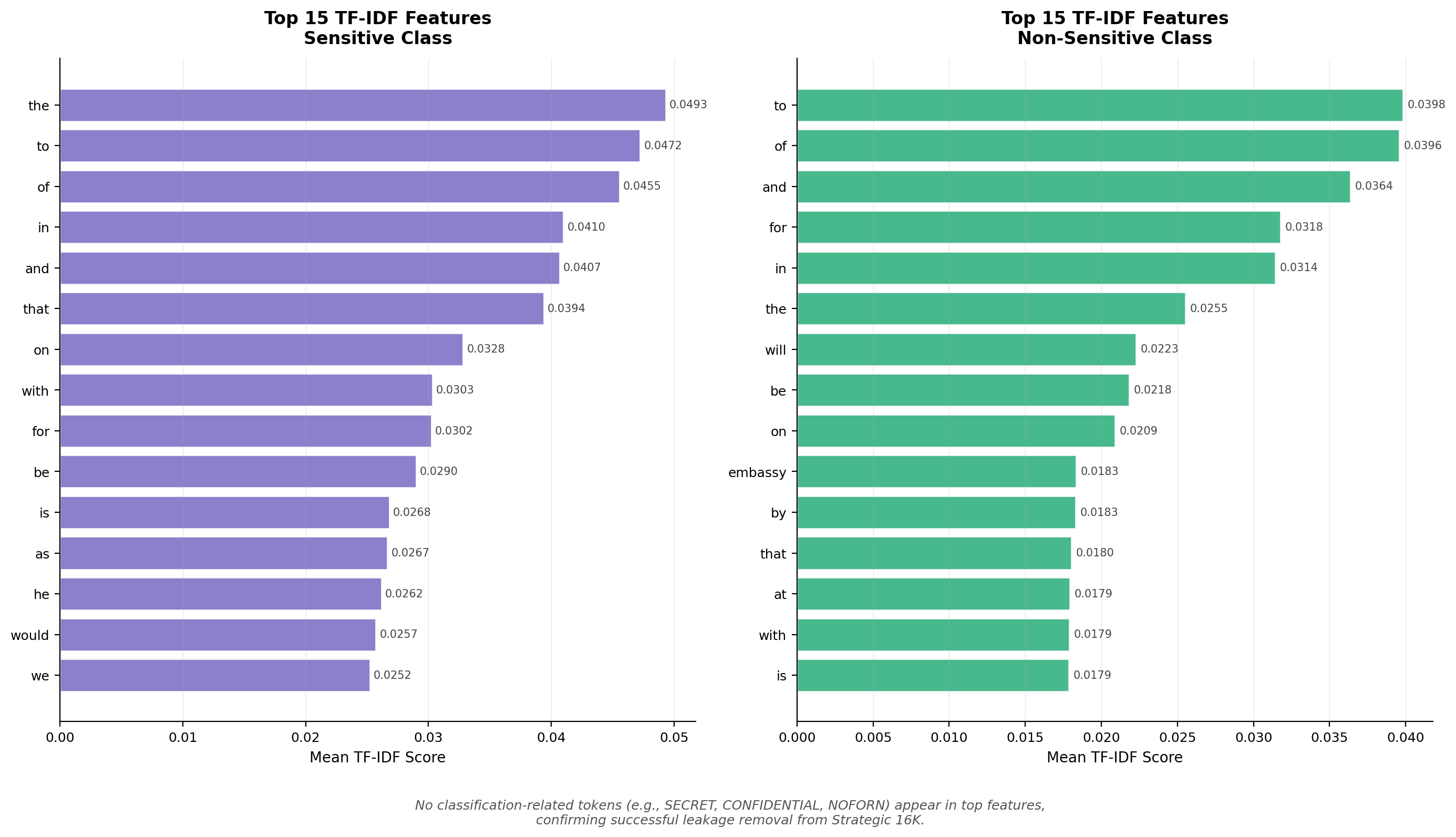}
  \caption{Top 15 TF-IDF features for Sensitive and Non-Sensitive
  classes on Strategic 16K. Both feature sets consist entirely of
  generic English vocabulary with no classification-related tokens,
  providing empirical confirmation of successful leakage removal.}
  \label{fig:featureaudit}
\end{figure}

To provide empirical confirmation of successful leakage removal,
the top TF-IDF features were examined for both classes on the
cleaned corpus. Fig.~\ref{fig:featureaudit} shows the top 15
features per class. Both feature sets consist entirely of generic
English vocabulary; no classification-related tokens such as
SECRET, CONFIDENTIAL, or NOFORN appear among the highest-weighted
features of either class. This confirms that classifiers trained
on Strategic 16K must rely on content-based signals rather than
surface classification artefacts.

\begin{table}[h]
\caption{Impact of label leakage on TF-IDF with Logistic Regression performance.}
\label{tab:leakage}
\centering
\begin{tabular}{lcc}
\toprule
\textbf{Dataset} & \textbf{Accuracy} & \textbf{F1} \\
\midrule
Raw PlusD (Leaky) & 99.0\% & 99.0\% \\
Strategic 16K (Cleaned) & 86.26\% & 86.83\% \\
\bottomrule
\end{tabular}
\end{table}

The severity of this leakage problem was confirmed empirically.
As shown in Table~\ref{tab:leakage}, TF-IDF with Logistic
Regression achieves nearly perfect performance on the raw corpus,
but its performance decreases substantially after leakage removal.
This indicates that the inflated performance on the raw dataset is
primarily attributable to residual classification artefacts rather
than genuine semantic sensitivity understanding.

To further validate the corpus construction, a length-only baseline
was evaluated in which document length alone was used as the
classification signal. This baseline achieved substantially lower
accuracy than all evaluated models, confirming that the length
difference between Sensitive and Non-Sensitive documents in
Strategic 16K cannot serve as a reliable proxy for sensitivity. The
corpus therefore forces models to learn from content rather than
structural shortcuts, and the performance reported in this benchmark
reflects genuine sensitivity discrimination.

\subsection{Strategic 16K Benchmark}

Following leakage removal, \textit{Strategic 16K} was assembled by
applying a sensitive-focused sampling strategy that increases model
exposure to the diversity of sensitive document patterns during
training. As shown in Table~\ref{tab:dataset}, a total of 16,000
documents with a near-balanced class distribution (Sensitive 52.4\%,
Non-Sensitive 47.6\%) have been included in the corpus. The
near-balanced distribution avoids having one class being structurally
privileged when applied to cross-validation evaluation.
The corpus was constructed via random sampling from the
leakage-cleaned document pool, with class counts adjusted to
achieve near-balance. The complete Strategic 16K dataset is
publicly available to support reproducibility and future
benchmarking at: {\small\url{https://drive.google.com/file/d/11-a5QeBklbrefKHdlcwoKPAZQ6mo119w/view?usp=drive_link}}

Deduplication was done at the document level before splitting, so
that there is no overlap in the documents contained in different
folds. Stratified sampling was used for fold construction for each
cable individually, but no constraints were applied to grouping at
the cable level; each cable in PlusD is a separate document and a
separate classification label is used for each one. All the fold
assignment is fully reproducible by using a fixed random seed of 42.

\begin{figure}[h]
  \centering
  \includegraphics[width=\columnwidth]{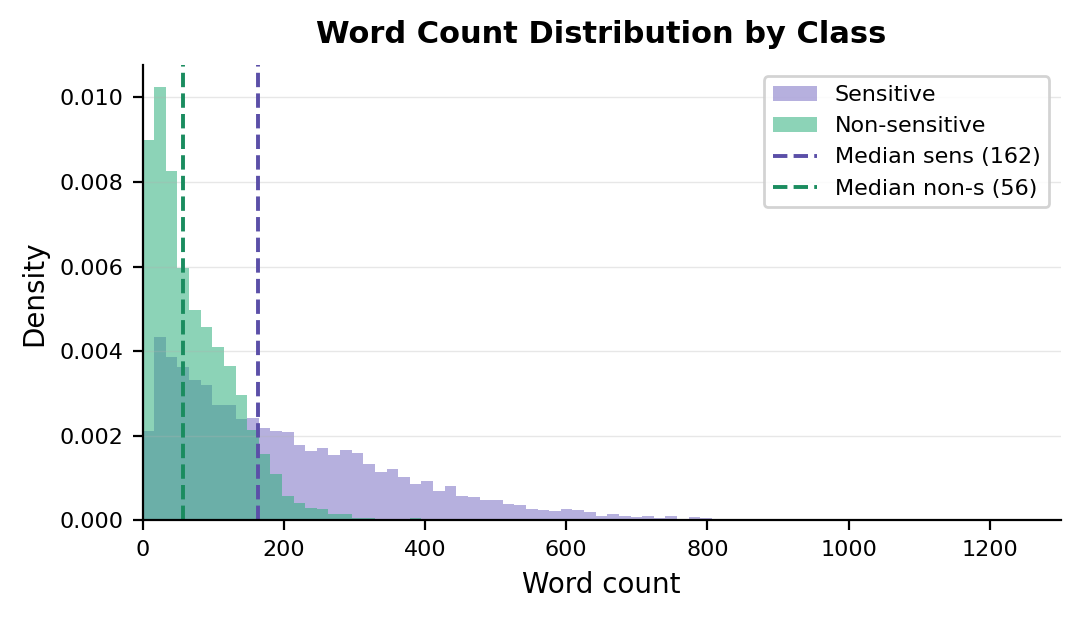}
  \caption{Word count distribution by class. Sensitive documents
  (median 276 words) have a significantly longer and wider
  distribution than Non-Sensitive documents (median 64 words).}
  \label{fig:wordcount}
\end{figure}

\begin{figure}[h]
  \centering
  \includegraphics[width=\columnwidth]{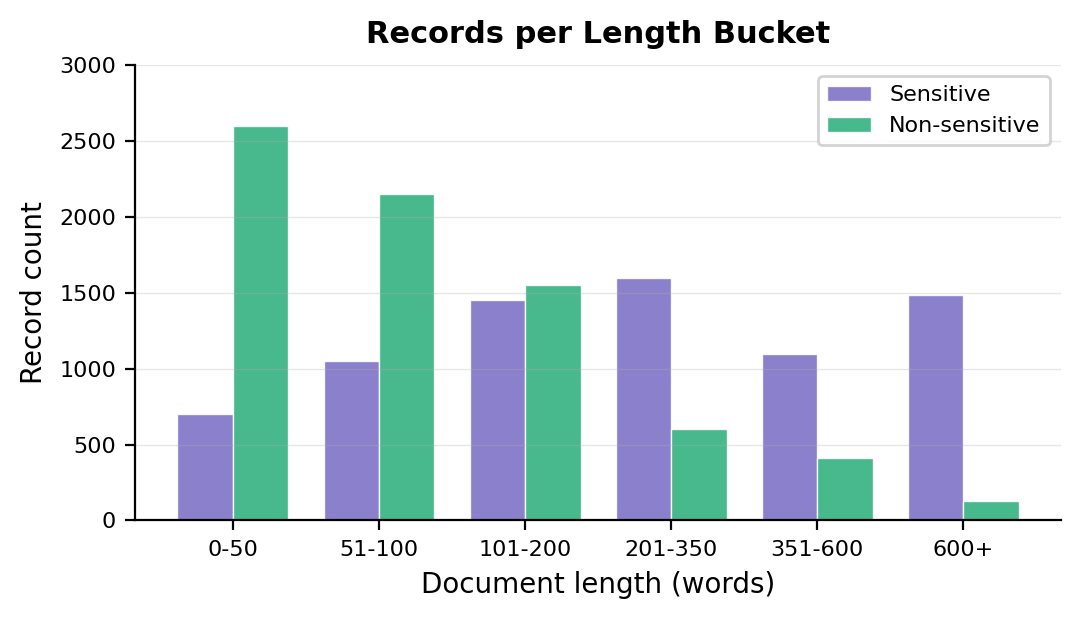}
  \caption{Records per document length bucket. Sensitive documents
  dominate longer buckets while Non-Sensitive documents concentrate
  in the 0--100 word range.}
  \label{fig:lengthbucket}
\end{figure}

The median of the Sensitive documents is 276 words, and the median
of the Non-Sensitive documents is 64 words.
Fig.~\ref{fig:wordcount} illustrates the word count distribution
across both classes. Simple length-thresholding is not a reliable
classification strategy due to the variety of sensitive document
lengths, from short cables to documents longer than 1,400 words.
Fig.~\ref{fig:lengthbucket} also illustrates how sensitive documents
dominate longer length buckets, further establishing the wide
coverage of the corpus in terms of length.

\begin{table}[h]
\caption{Binary Label Mapping and Strategic 16K Composition}
\label{tab:dataset}
\centering
\renewcommand{\arraystretch}{1.2}
\begin{tabular}{lcc}
\toprule
\textbf{Original Label} & \textbf{Binary Class} & \textbf{Count} \\
\midrule
UNCLASSIFIED             & Non-Sensitive & \multirow{2}{*}{7,613} \\
UNCLASSIFIED//FOUO       & Non-Sensitive & \\
LIMITED OFFICIAL USE     & Sensitive     & \multirow{5}{*}{8,387} \\
CONFIDENTIAL             & Sensitive     & \\
CONFIDENTIAL//NOFORN     & Sensitive     & \\
SECRET                   & Sensitive     & \\
SECRET//NOFORN           & Sensitive     & \\
\midrule
\textbf{Total}           & \textbf{2 classes} & \textbf{16,000} \\
\bottomrule
\end{tabular}
\end{table}

\section{Models and Experimental Setup}

\subsection{Evaluation Protocol}

All six models were evaluated using 5-fold stratified
cross-validation on Strategic 16K. Stratified splitting ensures that
the 52.4\%/47.6\% class ratio is preserved identically across all
folds. A fixed random seed of 42 was used throughout to ensure full
reproducibility. Weighted F1-score and accuracy are reported as the
primary evaluation metrics for all models. Weighted F1 is preferred
as it accounts for the slight class imbalance and jointly captures
precision and recall.

In security-sensitive classification tasks, the cost of a false
negative, that is, classifying a sensitive document as
non-sensitive, substantially exceeds the cost of a false positive.
An undetected sensitive document may be routed, stored, or
transmitted without appropriate access controls, with potentially
severe operational and regulatory consequences. Accordingly,
sensitive class recall is treated as a secondary evaluation
criterion alongside weighted F1, as it directly quantifies the
model's ability to avoid missed detections of sensitive content.

\subsection{Classical TF-IDF Models}

Three classical models were evaluated, each using TF-IDF feature
representations with sublinear term frequency scaling.
\textbf{Logistic Regression} applies L2 regularization to a linear
decision boundary. Hyperparameters were tuned via RandomizedSearchCV
with 25 candidates and 5-fold inner cross-validation, selecting
$C = 25.13$ with balanced class weights. \textbf{Linear SVM} finds
the maximum-margin separating hyperplane in TF-IDF
space~\cite{joachims1998}; best parameters selected $C = 14.53$
with balanced class weights. A convergence warning from the
liblinear solver indicates that the optimisation did not fully
converge, meaning reported results represent a conservative lower
bound. \textbf{Multinomial Naive Bayes} applies the conditional
independence assumption with Laplace
smoothing~\cite{mccallum1998}; best $\alpha = 0.021$, fit
prior\,=\,False.

\subsection{Transformer-Based Models}

Three transformer architectures were fine-tuned end-to-end under
5-fold stratified cross-validation with maximum token length 256.
\textbf{BERT} (bert-base-uncased)~\cite{devlin2019} is pre-trained
on masked language modelling (MLM), randomly masking 15\% of tokens
and training the model to reconstruct them using bidirectional
context; a linear classification head is attached to the
\texttt{[CLS]} token. \textbf{RoBERTa} extends BERT with dynamic
masking, removal of the next-sentence prediction objective, and
training on substantially larger data, modifications intended to
produce stronger general-purpose representations.
\textbf{ELECTRA}~\cite{clark2020} uses a replaced token detection
(RTD) objective: a generator replaces some tokens with plausible
alternatives, and the discriminator is trained to identify which
tokens were replaced. RTD trains on every input token rather than
only the masked 15\% subset, yielding greater sample efficiency and
representations grounded in token plausibility rather than
surface-form reconstruction.
Encoder based transformers have also inspired the creation of modern Large Language Models (LLMs), which are large-scale pretrained models based on the transformer architecture and have achieved outstanding performance in many natural language understanding and generation tasks~\cite{minaee2025}.

\section{Results and Discussion}

\subsection{Overall Performance}

\begin{figure}[h]
  \centering
  \includegraphics[width=\columnwidth]{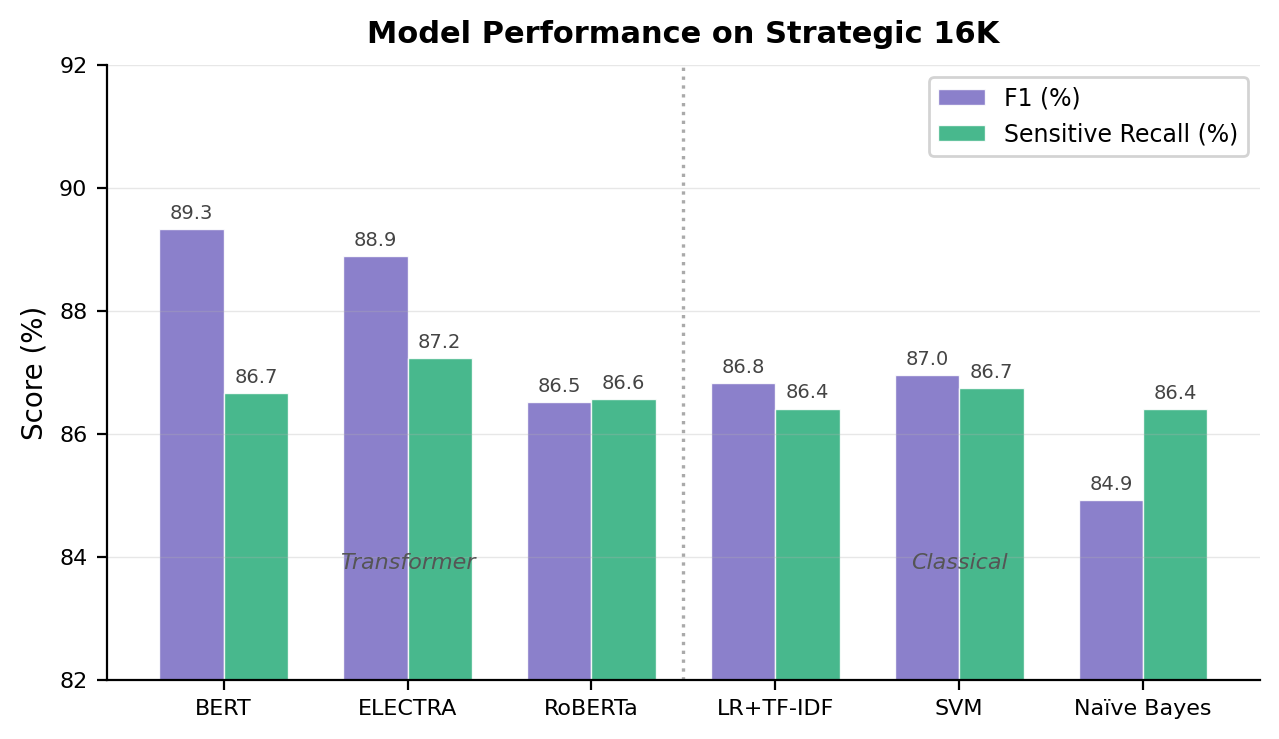}
  \caption{F1 scores for all six models on Strategic 16K. A clear
  performance gap separates the transformer tier (BERT, ELECTRA,
  RoBERTa) from the classical tier (LR, SVM, Na\"{i}ve Bayes).}
  \label{fig:results}
\end{figure}

Table~\ref{tab:results} presents the complete 5-fold
cross-validation results for all six models on Strategic 16K. BERT
achieves the strongest performance across both accuracy and F1.
ELECTRA trails BERT by 2.25\,pp on accuracy and 2.39\,pp on F1,
while remaining clearly ahead of the classical tier. The three
classical TF-IDF models form a distinct lower performance tier, with
SVM leading marginally on F1 and recall. RoBERTa occupies an
intermediate position, underperforming both BERT and ELECTRA.
Fig.~\ref{fig:results} visualises F1 and sensitive recall across
all six models, with a clear separation between the transformer and
classical tiers.

\begin{table}[h]
\caption{5-Fold CV Results on Strategic 16K}
\label{tab:results}
\centering
\renewcommand{\arraystretch}{1.2}
\resizebox{\columnwidth}{!}{%
\begin{tabular}{lccccc}
\toprule
\textbf{Model} & \textbf{Acc.} & \textbf{F1} & \textbf{S-Recall} & \textbf{S-Prec.} \\
\midrule
\textbf{BERT}   & \textbf{89.14\%} & \textbf{89.33\%} & 86.66\%          & \textbf{92.21\%} \\
ELECTRA         & 88.57\%          & 88.90\%          & \textbf{87.24\%} & 90.70\%          \\
RoBERTa         & 85.85\%          & 86.51\%          & 86.57\%          & 86.45\%          \\
LR + TF-IDF     & 86.26\%          & 86.83\%          & 86.41\%          & 87.25\%          \\
SVM             & 86.35\%          & 86.95\%          & 86.74\%          & 87.16\%          \\
Na\"{i}ve Bayes & 83.91\%          & 84.92\%          & 86.40\%          & 83.48\%          \\
\bottomrule
\multicolumn{5}{l}{\footnotesize S-Recall = Sensitive Recall, S-Prec = Sensitive Precision.}
\end{tabular}}
\end{table}

\subsection{Transformer vs.\ Classical Gap}

The results reveal a clear and consistent performance gap between
the two architectural families. BERT leads with 89.33\% F1 and
89.14\% accuracy. The gap between BERT and the best classical model
(SVM, 86.95\% F1) is 2.38\,pp on F1 and 2.79\,pp on accuracy,
widening to 6.37\,pp F1 against Naive Bayes. Paired t-tests confirm
that all transformer-vs-classical comparisons are statistically
significant: BERT vs.\ SVM ($t=15.19$, $p<0.001$, 95\% CI
[3.55\,pp, 5.14\,pp]); BERT vs.\ LR ($t=12.52$, $p<0.001$, 95\% CI
[3.48\,pp, 5.46\,pp]); BERT vs.\ Naive Bayes ($t=13.80$,
$p<0.001$, 95\% CI [5.09\,pp, 7.66\,pp]); and ELECTRA vs.\ SVM
($t=4.38$, $p=0.012$, 95\% CI [0.71\,pp, 3.18\,pp]). RoBERTa
(86.51\% F1) sits between the transformer and classical tiers,
performing comparably to the best classical models despite being a
full transformer architecture.

The transformer advantage comes from the fact that the fundamental capacity of representation is different. Transformer's self-attention architecture can be used to consider the bidirectional context between every word in the entire document, allowing for semantic context sensitive discrimination that is not sensitive to lexical frequency. The distinction between a diplomatic cable in a normal administrative setting and a diplomatic cable in a restricted operational setting is not captured reliably by a feature like TF-IDF, but can be captured by transformer attention via adjacent tokens.

\subsection{BERT vs.\ ELECTRA}

BERT outperforms ELECTRA by 0.57\,pp on accuracy and 0.43\,pp on
F1. A paired t-test across five fold-level F1 scores confirms that
BERT's advantage is statistically significant ($t = 8.20$,
$p = 0.0012$), with a mean F1 difference of 2.40\,pp and a 95\%
confidence interval of (1.58\,pp, 3.21\,pp).

The advantage is consistent with the nature of the task. BERT's
masked language modelling produces representations sensitive to a
broad range of semantic patterns in the diplomatic register.
ELECTRA's replaced token detection, which is based on token plausibility, rather than on complete reconstruction of the context in which tokens appear, might not be better suited to PlusD cables' specific vocabulary. Remarkably, ELECTRA outperforms in terms of sensitive recall (87.24\% vs.\ 86.66\%), which is useful in recall-sensitive deployments where minimizing false negatives is more critical than the overall F1.,
\subsection{Sensitive Class Recall Analysis}

From a security standpoint, sensitive class recall is the most
operationally critical metric, as it measures the proportion of
truly sensitive documents that are correctly identified. A notable
finding is that Naive Bayes achieves a sensitive recall of 86.40\%,
which is comparable to Logistic Regression (86.41\%) and SVM
(86.74\%), despite trailing both models substantially on F1 and
accuracy. This indicates that Naive Bayes, while less precise, is
relatively conservative in classifying documents as non-sensitive.
BERT achieves the strongest overall balance, with 89.33\% F1 and
86.66\% sensitive recall, making it the most reliable model for
deployment where both coverage and precision are required.

\subsection{Efficiency Tradeoff: Classical vs.\ Transformer}

When it comes to the classical models, SVM has the best performance (F1\,=\,86.95\%) and needs only standard CPU resources without any fine-tuning of the GPU. Logistic Regression is a close second (F1\,=\,86.83\%) and slightly quicker to tune (362.6 seconds as opposed to SVM's 382.2 seconds). TF-IDF-based models are extremely practical to deploy for organizations with limited computational resources, where the latency to inference is critical or when one wants to update the models quickly.

\section{Conclusion}

This paper presented a systematic benchmark evaluation of six
document sensitivity classification models on Strategic 16K, a
16,000-document leakage-controlled corpus constructed from WikiLeaks
PlusD diplomatic cables. We documented a three-category leakage
removal protocol targeting inline paragraph markers, embedded
classification phrases, and distribution notices, artefacts present
in raw WikiLeaks PlusD cables that prior studies have not explicitly
addressed, and demonstrated that removing them is essential for
producing honest, content-grounded performance estimates.

Among the six evaluated architectures, BERT achieves the strongest
and most stable performance (Accuracy\,=\,89.14\%, F1\,=\,89.33\%,
$\pm$0.46\% F1 std). ELECTRA performs competitively
(F1\,=\,88.90\%) and may be preferable in resource-constrained
settings given its parameter efficiency. Among classical models, SVM
achieves the highest F1 (86.95\%) and sensitive recall (86.74\%),
while Logistic Regression offers a comparable and slightly faster
alternative. RoBERTa achieves 86.51\% F1 but underperforms both
transformer peers, warranting dedicated hyperparameter
investigation. The sensitive class recall analysis confirms that
transformer models offer the best balance between coverage and
precision for security-critical deployment.

\subsection*{Limitations and Future Directions}

While BERT achieves the strongest benchmark performance, a
monolithic model has structural limitations in security-critical
deployment. Sensitivity is not determined by a single signal type:
a document may be sensitive because of who is mentioned, what topic
it covers, or how information is phrased. A single shared attention
mechanism cannot isolate, prioritise, or explain these signals
independently, and any policy change requires full model retraining.

Future work will address these limitations through a Multi-Agent
System (MAS) architecture in which specialised agents handle
distinct evidence categories: a \textit{semantic content agent}
evaluates topical sensitivity; a \textit{named entity agent} detects
references to restricted personnel, facilities, and programmes; and
a \textit{structural agent} evaluates subject lines and routing
cues. A learned fusion mechanism combines agent confidence estimates
to produce the final decision. This design provides built-in
explainability, since the fusion output can be decomposed to show
which agent drove each decision, and supports targeted updatability
when classification policies change. Explainable AI mechanisms
including LIME and SHAP will additionally be integrated for
per-document rationales, and the best-performing model will be
fine-tuned on real organizational document collections for
institutional domain adaptation. The Strategic 16K benchmark
documented here provides the controlled evaluation foundation
against which this proposed architecture will be measured.

\end{document}